# Agile and versatile bipedal robot tracking control through reinforcement learning

Jiayi Li, Linqi Ye, Yi. Cheng, Houde Liu*, and Bin Liang

*Abstract*—The remarkable athletic intelligence displayed by humans in complex dynamic movements such as dancing and gymnastics suggests that the balance mechanism in biological beings is decoupled from specific movement patterns. This decoupling allows for the execution of both learned and unlearned movements under certain constraints while maintaining balance through minor whole-body coordination. To replicate this balance ability and body agility, this paper proposes a versatile controller for bipedal robots. This controller achieves ankle and body trajectory tracking across a wide range of gaits using a single small-scale neural network, which is based on a model-based IK solver and reinforcement learning. We consider a single step as the smallest control unit and design a universally applicable control input form suitable for any single-step variation. Highly flexible gait control can be achieved by combining these minimal control units with high-level policy through our extensible control interface. To enhance the trajectory tracking capability of our controller, we utilize a three-stage training curriculum. After training, the robot can move freely between target footholds at varying distances and heights. The robot can also maintain static balance without repeated stepping to adjust posture. Finally, we evaluate the tracking accuracy of our controller on various bipedal tasks, and the effectiveness of our control framework is verified in the simulation environment.

## I. INTRODUCTION

Reinforcement learning (RL) methods for controlling legged robots have become a widely recognized research field with remarkable achievements in different application scenarios [1]-[4]. The goal of locomotion control can generally be summarized in five aspects: robustness, power, precision, lifelikeness, and intelligence. Adapting to modeling mismatch and environmental interference is the primary goal of legged robot motion control problem. Several sim-to-real technique for RL have been demonstrated to enhance the robustness of control strategies by bridging the "reality gap" between strategies trained in simulation and their real-world counterparts. Domain randomization [5][8] is one of the typical methods that significantly improve the adaptivity of policies by randomizing the dynamics of the simulator during training. Privileged learning [9][10] provides a way for policies to gain rich experience in traversing intricate terrains learned from privileged information available only in simulation, which can be distilled via supervised learning. The motor adaptation algorithm [11] designs an adaptation module that takes fully utilizes proprioceptive historical state data to predict the extrinsic and solve the real-time online adaptation problem. Powerful and precise are a group of challenging goal for RL, requiring strong driving capability as well as accurate and comprehensive control ability. The ostrich-like bipedal robot Cassie ran 100 meters in only 24.73 seconds using an RL policy. Li et al. [12] present an RL framework for Cassie to achieve robust and versatile dynamic jumps, such as jumping to different specified locations and directions. To synthesize graceful and life-like locomotion, imitation learning, and generative adversarial networks (GANs) are introduced to imitate the behaviors of biological beings using large amounts of processed biological movement data for training[13][14].

With the rapid development in robotics and its related fields, including body locomotion control, computer vision (CV), natural language processing (NLP), and simultaneous localization and mapping (SLAM), the pursuit of robot intelligence capable of integrating multidisciplinary research findings is intensifying. Multitasking and autonomous decision-making capabilities are two key aspects of robotic control intelligence. A versatile robot capable of autonomously completing various tasks based on instructions and environmental conditions is desired. Several hierarchical learning frameworks [15][18] have been devised to decouple end-to-end learning across multiple levels and allow different levels to utilize different resolutions of time. High-level strategies achieve multitask control and seamless transitions by utilizing latent variables to invoke reusable and composable lower-level strategies. Brohan et al. [19][20] demonstrate the generalization capabilities of robotics by transferring knowledge from large, diverse, task-agnostic datasets to modern machine learning models. Huang et al. [21] extract actionable knowledge from large language models (LLMs) to synthesize robot trajectories through a dense sequence of 6-DoF end-effector waypoints. They leverage the code-writing capabilities of LLMs to interact with a vision-language model (VLM) which can ground the knowledge into 3D observation space, and showcase the ability to perform a

* Research supported by the National Natural Science Foundation of China under grants No. 62003188 and No.92248304.

J. Li, Y. Cheng and H. Liu are with the Center for Artificial Intelligence and Robotics, Tsinghua Shenzhen International Graduate School, Tsinghua University, 518055 Shenzhen, China. H. Liu is the corresponding author. (e-mail: lijiayi21@mails.tsinghua.edu.cn; chenge9191@163.com; liu.hd@sz.tsinghua.edu.cn).

L. Ye is with the Institute of Artificial Intelligence, Collaborative Innovation Center for the Marine Artificial Intelligence, Shanghai University, 200444 Shanghai, China.(e-mail: yelinqi@shu.edu.cn).

B. Liang is with the Navigation and Control Research Center, Department of Automation, Tsinghua University, 100084 Beijing, China (e-mail: bliang@tsinghua.edu.cn).

wide variety of everyday manipulation tasks specified in free-form natural language on a robotic arm.

Undoubtedly, synthesizing interdisciplinary knowledge is crucial for developing highly intelligent robot control algorithms. However, unlike the end-to-end control methods demonstrated in [19][21] which are primarily validated on mechanical arms, controlling complex floating base robot represented by legged robots can be much more challenging due to their stability being closely related to their physical structure. We suggest that adopting a low-level approach to robotic embodied intelligence is key to bridging motion control for specific robots and the high-level large models. In this scenario, the same high-level controller can function on different robots with their own local low-level controllers and achieve zero-shot generalization to new tasks.

The objective of this paper is to design a versatile bipedal robot control method that is open-ended and task-independent with an intuitive and interpretable control interface. The central contribution of this work is an agile and general bipedal robot tracking controller based on reinforcement learning, which endows the bipedal robot with embodied intelligence. Locomotion tasks are designed based on real-time robot trunk and ankle position and orientation, which can be precisely tracked unless compromises are essential for balance. We propose a control framework that combines model-based feedforward policy with learning-based feedback policy to generate highly customized locomotion for various motion patterns. Curriculum learning [22] is also utilized during training to mimic the progressive learning process of humans. The agility of motion, universal balancing ability, and excellent tracking proficiency of our controller are demonstrated through various common tasks in simulation.

## II. RELATED WORK

### A. Action Space Legged Control

Precise foot placement control has been implemented on both quadruped and bipedal robots using model-based and model-free methods. Previous works have demonstrated significant potential for integrating models and learning methods for location tracking in Cartesian space. For instance, the quadruped robot Max can execute highly challenging maneuvers such as rotating steps and single-pole jumps on poles of varying heights, relying on technologies such as robot vision positioning, terrain recognition, omnidirectional six-degree-of-freedom motion planning, and high-precision model predictive control [23]. Jenelten et al. [24] propose a hybrid control architecture that combines the advantages of a model-based planner and deep neural network policy simultaneously, achieving remarkable robustness, foot-placement accuracy, and terrain generalization. This framework serves as a successful example of combining model-based and model-free approaches to the legged robot control problem. Arm et al. [25] train a robust RL policy for a quadruped robot to track position target points for one foot. Various real-word tasks such as door opening, sample collection and pushing obstacles are demonstrated through teleoperation. Li et al. [12] present an RL framework for training a bipedal robot to accomplish jumps to specific locations and directions using a multi-stage training scheme. They utilize a policy that structurally encodes both the long-term input/output (IO) history and the short-term I/O history. Duan et al. [26] proposed a method to integrate knowledge of the legged robot system into neural networks, enabling task space action learning in terms of foot setpoints. They use a task space inverse dynamics controller to track the foot pitch, yaw and position generated by the RL policy. Conversely, in our previous work [27], we apply a reference signal as a feedforward instruction and use an RL policy to generate feedback signals. The effectiveness of this combination has been validated through a bunch of bipedal and quadrupedal motions.

### B. Versatile Multi-Task Control Frameworks

Multitask learning for robots working in various environment has become a new tough challenge after high accuracy and speed are widely realized among substantial single tasks. Cheng et al. [28] use a single neural net policy operating from a camera image with large scale reinforcement learning which can overcome imprecise sensing and actuation to output highly precise control behavior on quadrupedal parkour end-to-end. Brohan et al. [19][20] train a single, capable, large multi-task backbone model on data consisting of a large size of robotic arm tasks with human-provided demonstrations training data using a transformer architecture.

Hierarchical control structure is widely used to deal with multi-task control problem. Ito et al. [29] propose an easily scalable method in which multiple deep predictive learning (DPL) [30] modules calculate the prediction error in real time and the one with the minimum prediction error is automatically executed. Only the competing part of each module is required to be designed artificially, thus it's easy to add or delete the modules of different tasks. A unified reward is used for different parkour cases, and the robot is finally able to long jump, high jump, run over tilted ramps, and even walk on just front two legs. Cheng et al. [31] proposed a skill learning and composition framework in which a behavior tree that encodes a high-level task hierarchy from one clean expert demonstration is learned to compose low level skills that are successfully transferred to the real world via online adaptation. Peng et al. [14] achieve stylized physics-based virtual humanoid character control using adversarial motion priors. They propose an adversarial method for learning general motion priors from large unstructured datasets. Their approach does not necessitate synchronization between the policy and reference motion because the adversarial discriminator is trained using the dataset consisting all the motion priors of multiple tasks, thus composition of disparate skills emerges automatically from the motion prior without any high-level planner. Peng et al. [2] further extended their research to a great larger scale data-driven framework which combines techniques from adversarial imitation learning and unsupervised reinforcement learning to train a low-level latent variable model synthesized by a high-level policy to produce behaviors like that in the dataset.

While these algorithms can generate very fluid movements and switch between tasks, we suggest that these control frameworks fall short in adding new, highly definable movement patterns. Once policies are trained, it is difficult to adjust locomotion in action space based on interpretable instructions anymore, and this is what we will focus on in this paper.

## III. CONTROL STRUCTURE

### A. Overview

Model-based methods and model-free reinforcement learning have been broadly applied to tackling bipedal locomotion. To accomplish complex motion tasks, the former typically requires establishing dynamic models that make a trade-off between complexity and accuracy, as well as manually designed control structures for specific tasks. The latter exhibits superior generalization and robustness while lacking some interpretability in its end-to-end control structure. Inspired by the ideal of combining these two approaches [32]-[34], we extend it to a general bipedal robot trajectory tracking control problem for both robot trunk and ankles. The framework of our control method is shown in Fig.1.

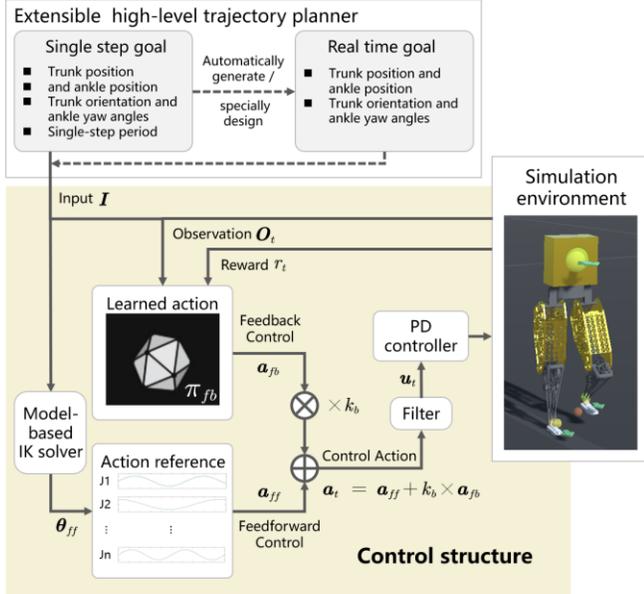

Figure 1. Overview of the control structure

We use an extensible high-level trajectory planner to generate control trajectories according to different task as input, which is fed to both a model-based IK solver to generate feedforward signals and a neural network to generate feedback signal. We simplify the algorithm's manual design complexity significantly by using simple inverse kinematics instead of inverse dynamics models. These two signals are combined through weighted summation and filtering to generate target positions for each joint. Finally, a PD controller is used to control the robot.

### B. Robot model and kinematics

In this paper, we use a modified 12-joint Ranger Max robot model [35], which differs from the original robot in several aspects. Each leg has three hip joints, one knee joints and two ankle joints. The general details of the robot model are shown in Table 1.

TABLE I. DETAILS OF THE ROBOT MODEL

| Link | Physical parameters | |
|---|---|---|
| | Size | Mass |
| Trunk | 0.3m×0.3m×0.15m | 20kg |
| Thigh | 0.4m (length) | 6.4kg |
| Shank | 0.35m (length) | 2kg |
| Foot | 0.16m (length) | 1kg |

We first define the concept of controllable nodes: controllable nodes are the coordinates on the robot whose position and orientation can be set as control target. In legged control tasks, the most concerned points of the robot are the trunk center, the left ankle, and the right ankle while the position and orientation of the knee and hip joints are always ignored. Thus, we identify the trunk node $n_T$, the left ankle node $n_L$, and the right ankle node $n_R$ as shown in Fig. 2.

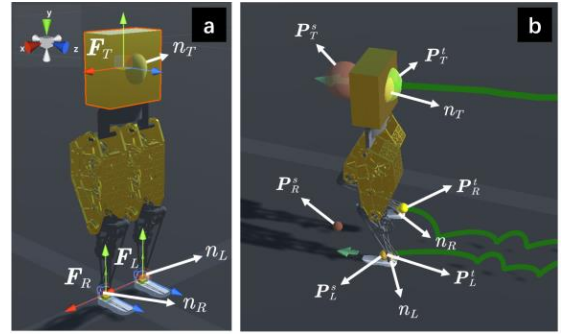

Figure 2. Controllable nodes of the robot (a) Controllable nodes and their coordinate system (b) Task-step target position and real-time target position of the controllable nodes

We establish the robot base coordinate system $F_B$ at $n_T$, the left ankle coordinate system $F_L$ at $n_L$, and the right ankle coordinate system $F_R$ at $n_R$ as shown in Fig. 2. Since the pitch and roll angles of the robot's feet during walking are typically adjusted based on the environment and the robot's own state, we only consider the yaw angle of the feet as the control target. However, to prevent the leg joint solving from becoming a complex redundant degree of freedom problem, we assume that the feet always remain horizontal with respect to the trunk's transverse plane to provide a feedforward signal

$$|F_{LY} \times F_{BY}|^2 + |F_{RY} \times F_{BY}|^2 = 0 \quad (1)$$

where $F_{.Y}$ is the Y-axis of coordinate system $F_.$. Then the feedforward joint angles can be obtained by solving the inverse kinematics equations with the positions and orientations of the three controllable nodes, constrained by the assumption of horizontal feet

$$\boldsymbol{\theta} = ikine(\boldsymbol{P}_L, \psi_L, \boldsymbol{P}_R, \psi_R) \quad (2)$$

where $\boldsymbol{\theta}$ is the vector of all the joint angles. $\boldsymbol{P}_L$ and $\boldsymbol{P}_R$ are the position of $n_L$ and $n_R$. $\psi_L$ and $\psi_R$ are the yaw angles of $n_L$ and $n_R$ respectively. Then variables mentioned above are represented in the robot base coordinate system $\boldsymbol{F}_B$. We prevent the issue of unsolvable inverse kinematics function by preprocessing the controllable nodes' positions of the ankles that are outside the robot's workspace, moving them closer to the interior of the workspace..

*C. Task description*

We define the movement of the foot from leaving the ground to returning to the ground as a task step. In our control framework, all the tasks are described as a series of task steps. Each task step consists of the trajectories of the three controllable nodes in a custom task-step period. The position and orientation of the controllable nodes at the end of a task-step period represent the target state of the robot after the completion of the task step. Within a task-step period, different gaits can be generated by designing smooth trajectories of each node about time. In this paper, we design a high-level trajectory planner to generate intermediate trajectories for certain end state of a task step. The fourth-order Bezier curves are employed to plan smooth ankle trajectories whose shape can be determined by the control points of the Bezier curve, allowing us the flexibility to adjust the shape of the ankle trajectories as desired. For the trunk nodes $n_T$, we typically provide a uniform velocity and angular velocity trajectory from the start to the end state of the task step. Note that this high-level planner is extensible if other kinds of trajectories are needed for complex tasks, and we have successfully experimented with using higher-order Bezier curves and trigonometric curves.

To make our control method widely applicable to various tasks, we have defined several basic task steps for training. By combining them, the robot can achieve a variety of task-decoupled changes in body state. The basic task steps are as shown in Table. 2.

TABLE II. BASIC TASK STEP INTRODUCTION

|   | Task step | Task step introduction |
|---|---|---|
| 1 | Standing still | The robot stands still in random posture with different trunk and ankle nodes position and orientation. |
| 2 | Squat | The robot adjusts its trunk position and orientation without moving its feet. |
| 3 | Walking backward/forward | The robot moves forward and backward in a straight line with different trunk heights. |
| 4 | Sidle | The robot moves left and right in a straight line with different trunk heights. |
| 5 | Turn | The robot turns clockwise or counterclockwise with a random radius. |
| 6 | Walking up and down stairs | The robot goes up and down steps of different heights and lengths. |

For the first two basic tasks, the feet never leave the ground so that the task-step period is 0 seconds. For the other tasks, the task-step period is a random value between 0.4 seconds and 0.7 seconds. The shapes of ankle trajectories are controlled by several Bezier control points which vary randomly within a certain range. In our control structure, all these basic task steps take the form of a unified control input

$$\begin{aligned}
\boldsymbol{I} &= [\,\boldsymbol{I}^s \ \ \boldsymbol{I}^t \ \ \boldsymbol{T}\,] \\
\boldsymbol{I}^s &= [\,\boldsymbol{P}_T^s \ \ \boldsymbol{P}_L^s \ \ \boldsymbol{P}_R^s \ \ \boldsymbol{R}_T^s \ \ \psi_L^s \ \ \psi_R^s\,] \\
\boldsymbol{I}^t &= [\,\boldsymbol{P}_T^t \ \ \boldsymbol{P}_L^t \ \ \boldsymbol{P}_R^t \ \ \boldsymbol{R}_T^t \ \ \psi_L^t \ \ \psi_R^t\,] \\
\boldsymbol{T} &= [\,T_L^t \ \ T_R^t\,]
\end{aligned} \quad (3)$$

where the superscript $s$ describes the final state of a task step and $t$ describes the real time state. $\boldsymbol{P}_T$, $\boldsymbol{P}_L$, $\boldsymbol{P}_R$ are the target position of $n_T$, $n_L$ and $n_R$ as shown in Fig. 2(b). $\boldsymbol{R}_T$ is the target orientation of $n_T$. $\psi_L$ and $\psi_R$ are the target yaw angles of $n_L$ and $n_R$. The elements of $\boldsymbol{I}^s$ and $\boldsymbol{I}^t$ are all presented in $\boldsymbol{F}_B$. $\boldsymbol{T}$ acts as a timer, where $T_L^t$ and $T_R^t$ represent how long the left and right foot are expected to touch the ground respectively. For the basic task standing still and squat in which the feet are always in contact with the ground, $\boldsymbol{T}$ is always a zero vector..

*D. Instruction learning*

In our previous work, we refer to a learning method that combines feedforward and feedback as "instruction learning" [27], which is inspired by the human learning process and is highly efficient, flexible, and versatile for robot motion learning. We design the control architecture of this paper based on the same approach and extend it to a more general framework. We pass the input in (3) to both a neural network and a model-based inverse kinematics (IK) solver, as shown in Fig. 1. The IK solver calculates the feedforward joint angles according to the real time target state of the controllable nodes using the inverse kinematics equations in (2). Then we normalize the values in $\boldsymbol{\theta}_{ff}$ to [-1,1] to balance the unit

$$\boldsymbol{\theta}_{ff} = ikine(\boldsymbol{P}_L^t, \psi_L^t, \boldsymbol{P}_R^t, \psi_R^t) \quad (4)$$

$$\boldsymbol{a}_{ff} = normalization(\boldsymbol{\theta}_{ff}) \quad (5)$$

The output vector of the feedback network $\boldsymbol{a}_{fb}$ whose element is also limited in [-1,1] represents the feedback adjustment for each joint. The control action signal is a weighted sum of the feedforward and feedback signals

$$\boldsymbol{a}_t = \boldsymbol{a}_{ff} + k_b \times \boldsymbol{a}_{fb} \quad (6)$$

where $k_b$ is the feedback ratio. By adjusting $k_b$, the boundaries of the control action are specified as shown in Fig. 2. The influence of $k_b$ has been discussed in [27].

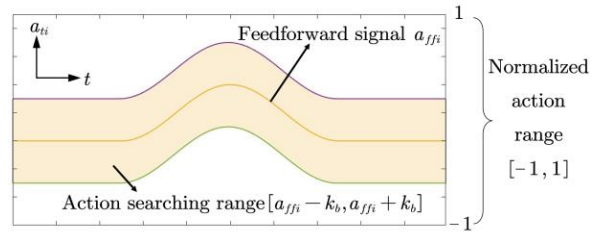

Figure 3. Illustration of action bounding

To prevent the robot joints from exceeding their physical limits, we clip the control signals to the range of [-1, 1] using the $calmp(\cdot, \min, \max)$ function. Additionally, to mitigate the risk of robot jitter caused by large differences in adjacent joint angle signals, we employ a simple first-order low-pass filter to filter the control signals.

$$u_t = calmp(\alpha_{filter} a_t + (1 - \alpha_{filter}) a_{t-1}, -1, 1) \quad (7)$$

where $\alpha_{filter}$ is the filtering coefficient and $a_{t-1}$ represents the action signal of the previous time step. Finally, $u_t$ is mapped to the range of joint angles for each joint and used to control the robot through the PD controller.

## IV. RL PROBLEM FORMULATION

To obtain a feedback policy that cooperates with the model-based feedforward IK-solver, we employ reinforcement learning to train the feedback policy. Proximal policy optimization (PPO) [36] is used to train the feedback policy $\pi_{fb}$ due to its stability, sample efficiency and strong adaptability to continuous action space.

### A. Hyperparameter and neural network

Some of the main training hyperparameters are shown in Table. 3. The values of learning rate, beta and epsilon varies in different training stages. We achieve finer adjustment of the network by gradually reducing their maximum values and choose a linear learning schedule. The feedback network has an actor-critic structure. The actor network is a multi-layer perceptron (MLP) with 3 hidden layers with 512 hidden units for each layer. The critic network is another MLP with 2 hidden layers with 128 hidden units for each layer.

TABLE III. TRAINING HYPERPARAMETERS

| Hyperparameter | Value (Stage1/2/3) |
|---|---|
| batch_size | 2048 |
| buffer_size | 20480 |
| learning_rate | 0.0003/0.0002/0.0001 |
| beta | 0.02/0.015/0.07 |
| epsilon | 0.02/0.015/0.07 |
| lambda | 0.95 |
| num_epoch | 3 |

### B. Observation and action

The observation of the learning strategy is designed as

$$O_t = [I_{error}^s \quad I_{error}^t \quad I_v^t \quad T \quad O_{state} \quad a_{ff} \quad a_{fb(t-1)}]$$
$$O_{state} = [g \quad v_T \quad w_T \quad J_P \quad J_V] \quad (8)$$

where $I_{error}^s$ is the error between the target task-step end state $I^s$ and the real controllable nodes state, and $I_{error}^t$ is the error between the target real-time state $I^t$ and the real controllable nodes state. $I_v^t$ is the derivative of $I^t$ with respect to time. $g$ represents for gravity. $v_T$ and $w_T$ are the velocity and angular velocity of $n_T$. All the observation elements above are measured in $F_B$. $J_P$ and $J_V$ are the joint angle position vector and the joint angle velocity vector. The action of the network $a_{fb}$ is a feedback vector that reflects what direction and how much adjustment joint angles should make based on the current robot target and state in addition to the feedforward signal. Each element in $a_{fb}$ is within [-1,1]. $a_{fb(t-1)}$ is the output vector of the feedback network at the previous time.

$$a_{fb} = [a_{fb1} \quad a_{fb2} \quad \cdots \quad a_{fb12}]$$
$$a_{fbi} \in [-1, 1], \ i = 1, 2, \cdots, 12 \quad (9)$$

$$a_{fb} = [a_{fb1} \quad a_{fb2} \quad \cdots \quad a_{fb12}], \ a_{fb.} \in [-1, 1] \quad (10)$$

### C. Reward design

Due to the introduction of the feedforward signal, the feedback network no longer needs to contain a large amount of information related to the robot model. It focuses more on fine-tuning the robot in the current state to achieve coordinated motion and tracking of controllable points. Thus, the task-specific signals are also included in the feedforward part, so we can reward the network in a uniform form which is task-independent

$$r_t = r_{live} + r_{error} + r_{fb} + r_{static} \quad (11)$$

where $r_{live}$ is a constant reward that encourage the robot from falling. $r_{error}$ represents the reward for encouraging the robot's controllable node states to be as close as possible to the real-time target states. $r_{error}$ has three parts corresponding to three controllable nodes

$$r_{error} = r_{Terror} + r_{Lerror} + r_{Rerror} \quad (12)$$

$r_{.error}$ is formulated exponentially and represents the state tracking error of controllable nodes $n_.$, the reward forms are as follow

$$r_{Terror} = w_{TP} e^{-\sigma_{TP} \| P_T^s \ominus P_T \|} + w_{TR} e^{-\sigma_{TR} \| R_T^s \ominus R_T \|} \quad (13)$$

$$r_{ierror} = w_{iP} e^{-\sigma_{iP} \| P_i^s \ominus P_i \|} + w_{i\psi} e^{-\sigma_{i\psi} \| \psi_i^s \ominus \psi_i \|}, i = L, R \quad (14)$$

$\ominus$ denotes the quaternion difference for trunk orientation, the yaw angle difference for ankle orientation and the vector difference otherwise. The ideal of trunk "soft" tracking introduced in [24] is employed here as the position and orientation of ankles are always much more important than that of the trunk. When necessary, the tracking error of trunk will be compromised for whole body balance maintenance and ankle tracking, thus we set the position error weight $w_{TP}$ and the orientation error weight $w_{TR}$ of the trunk to be only 0.2 times that of the ankle error weight $w_{iP}$ and $w_{i\psi}, i = L, R$. $r_{fb}$ is the reward to punish excessive feedback signals, which encourage the control policy to be more "lazy"

$$r_{fb} = w_{fb} \text{clamp}\left(\sum_{i=1}^{12} \sigma_{fbi} a_{fbi}, -1, 0\right) \quad (15)$$

The final reward $r_{static}$ is used to penalize the trembling of the robot and is only activated when the target velocities of all three controllable nodes are stationary.

$$r_{static} = \begin{cases} 0 & stationary\ task \\ r_{staticJV} + r_{staticJA} & otherwise \end{cases}$$

$$r_{staticJV} = w_{JV} \text{clamp}\left(\sum_{i=1}^{12} \sigma_{JVi} J_{Vi}, -1, 0\right) \quad (16)$$

$$r_{staticJA} = w_{JA} \text{clamp}\left(\sum_{i=1}^{12} \sigma_{JAi} J_{Ai}, -1, 0\right)$$

where $J_{Vi}$ and $J_{Ai}$ are the velocity and acceleration of the $i$th joint, $i = 1, 2, \cdots, 12$.

### D. Multi-stage training and episode design

The final goal of the training is to develop a control policy that enables the robot to flexibly execute a variety of single-step movements which is introduced as task-step in this paper, and complete diverse task by combining these single-step movements. To ensure the universality and diversity of tasks, it is necessary to consider different kinds of single-step variations, including changes in the position and orientation of the trunk and ankle nodes in both the horizontal plane and the vertical direction. We design a three-stage training curriculum to gradually increase the complexity of single-step movements as shown in Fig. 4.

We prepared unique pools of single-step movements for each training stage. During training, a random movement from the pool is selected for each episode. In the first training stage, we focus on learning motions on level ground, and we put the first 5 motions in Table. 2 into the movement pool. One training stage is divided into a few training lessons, where we gradually increase the complexity of ankle node target trajectories, such as increasing the step length and the leg lifting height. In the second stage, we add stair climbing into the movement loop. Since stairs can be regarded as obstacles during leg movement, the ankle trajectory tracking ability acquired in the first training stage serves as a preparatory skill for stair climbing actions in the second stage. We attempt to learn stair climbing from the beginning, only to find that the robot would continuously perform stationary stepping movements to avoid tripping or falling down stairs. At this point, the robot is capable of freely controlling its legs to perform various gaits. We proceed with a third-stage static balance training process. For all movements in the pool, we instruct the robot to maintain the state of the controllable nodes after completing a certain number of single-step movements, allowing the robot to maintain the posture obtained after each single-step movement. Through this training, we free the robot from the exhausting state of repeatedly stepping to adjust its body position to maintain balance.

To train the network more efficiently, episodes need to be terminated promptly when the robot's state deviates from the goal and there is little possibility of recovery. An episode terminates when the tilt angle of the robot trunk exceeds 60 degrees or when the episode duration reaches the maximum training time.

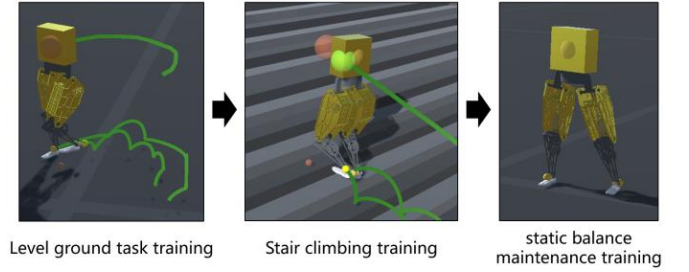

Figure 4. Three-state training

## V. SIMULATION RESULTS

We use Unity and ML-Agents for simulation and training to validate our methodology. The PhysX engine is adopted. The time step for each action is 0.005s, which indicates a control frequency of 200 Hz. The training is running on the CPU of a personal computer. To speed up the training, we use 50 copies of the agents for parallel training. The whole training process which has 300 million steps takes around 70 hours.

After the three-stage learning, the robot can achieve all the tasks listed in Fig. 2. We tested each type of them and recorded the errors for 10 seconds. The tracking errors for different tasks are shown in Fig.5 to Fig. 10 and the visualized trajectories are depicted in Fig. 11, where the green trajectories represent the nodes' target trajectories, and the yellow trajectories represents the nodes' actual trajectories. It can be noticed that the average position tracking error of both left and right ankles is within 5 centimeters or less, and the average orientation tracking error of ankles is within 5 degrees or less. The tracking error of the trunk trajectory is relatively large. Considering that the trunk tracking trajectory provided does not have an actual dynamic basis, this phenomenon is acceptable under the soft constraint condition that compromises trunk tracking for balance.

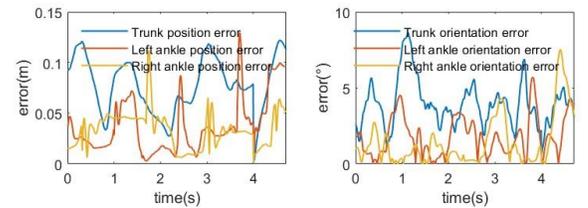

Figure 5. Tracking error of walking down stairs

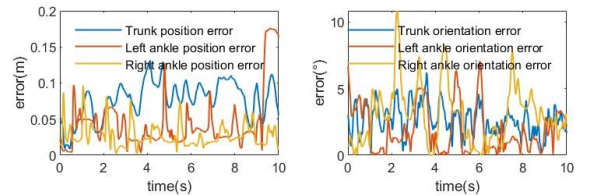

Figure 6. Tracking error of walking forward

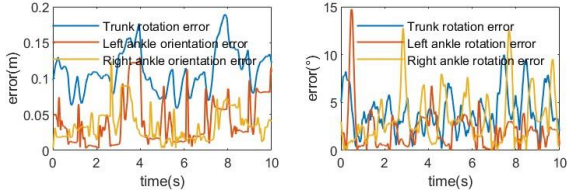

Figure 7.  Tracking error of walking backward

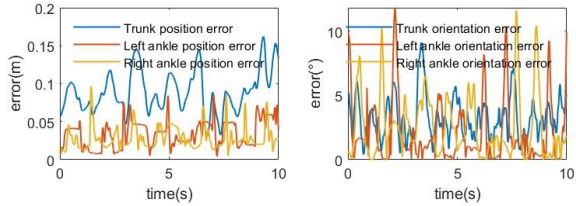

Figure 8.  Tracking error of sidle

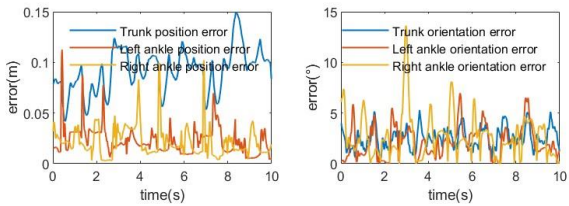

Figure 9.  Tracking error of turn

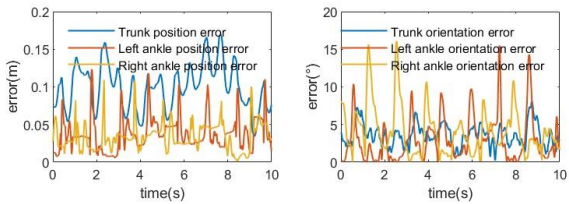

Figure 10.  Tracking error of walking up stairs

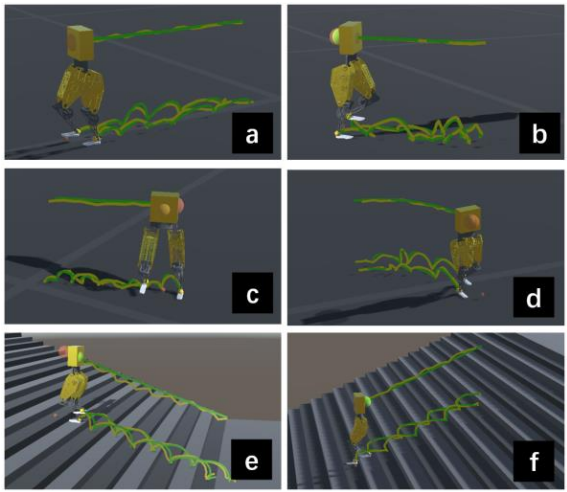

Figure 11.  Trajectories in different tasks

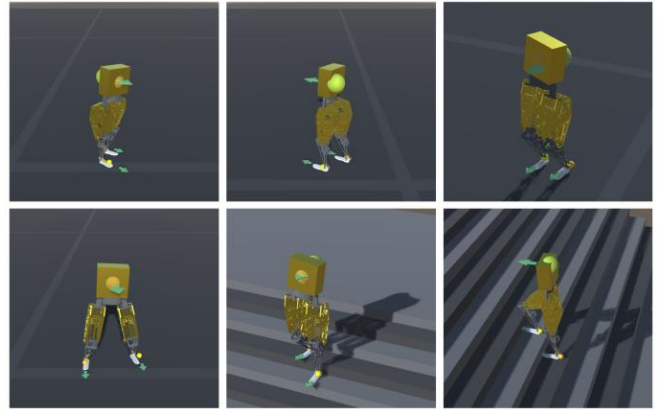

Figure 12.  Static balance maintainance

We also conducted experiments on the robot's static balance capability. For all movements in the task pool, we allow the robot to move multiple steps and then stop updating the target foothold. We find that the robot could maintain its current posture as shown in Fig. 12. So far, we have validated the effectiveness of our proposed framework in level ground locomotion, stair climbing, and static balance maintenance. All feedback motion control is achieved by a single small-scale network. Through designing the position of footholds, we also tested the obstacle avoidance capability of our controller, as shown in Fig. 13. This demonstrates that our universal controller can accomplish a variety of tasks through flexible combinations of single-step movements.

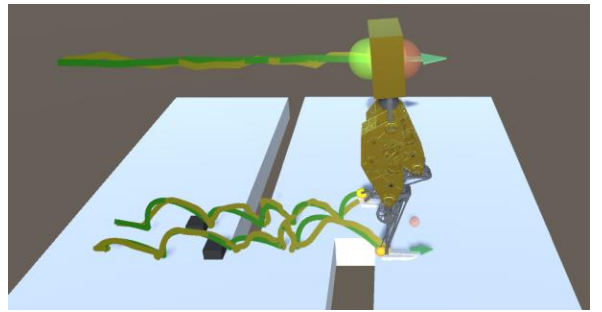

Figure 13.  Obstacle avoidance

## VI.  Conclusion and Future work

In this paper, we propose a method that combines both model-based and RL approaches for agile and versatile bipedal robot tracking control. We design a universal form of control interface, enabling the robot to accomplish various tasks through combinations of different single-step movements. Through a small-scale three-layer network and a simple IK solver, our controller has achieved excellent and general tracking capabilities for foot and body trajectories as well as static balance capabilities. The main contributions of this paper are: (1) Proposing a task-independent balance control approach that decouples robot motion control from task requirements, enabling the robot to achieve intrinsic balance capability solely based on its physical structure. (2) Providing a real-time precise ankle position tracking method, allowing for controlled leg movement height in robots, with interpretability. The effectiveness of our approach is validated through simulation experiments.

Thanks to the human-like properties of our control framework, we anticipate achieving more intricate movements, such as single-leg jumping, one-legged stance, and long jumping, among others, using the extensible high-level trajectory planner. Additionally, our controller's task inputs are defined by node states, which underscores its potential for human teleoperation of the bipedal robot via wearable sensor devices. Our immediate priority is to implement the algorithm on the physical humanoid robot system currently under construction. Subsequently, we will expand its application to additional tasks, culminating in the development of a teleoperation system for bipedal robots built upon this control framework.